\documentclass[12pt]{article}

\usepackage{amsmath,amsfonts,bm}

\def\eqref#1{equation~\ref{#1}}

\def\1{\bm{1}}

\DeclareMathAlphabet{\mathsfit}{\encodingdefault}{\sfdefault}{m}{sl}
\SetMathAlphabet{\mathsfit}{bold}{\encodingdefault}{\sfdefault}{bx}{n}

\usepackage[left=2.5cm,
right=2.5cm,
top=2.3cm,
bottom=2.3cm,
headheight=20pt,
headsep=10pt,
footskip=25pt,
letterpaper]{geometry}
\usepackage[utf8]{inputenc}
\usepackage[T1]{fontenc}
\usepackage[english]{babel}
\usepackage{amsmath,amsfonts,amssymb,thmtools,amsthm}
\usepackage{graphicx}
\usepackage{hyperref}
\usepackage{fancyhdr}
\usepackage{graphicx}
\usepackage{stfloats}
\usepackage{wrapfig}
\usepackage{epstopdf}
\usepackage{cleveref}
\usepackage{subfloat}
\usepackage{subcaption}
\usepackage{enumitem}
\usepackage{listings}
\usepackage{titlesec}
\usepackage{etoolbox}
\usepackage{setspace}
\usepackage{changepage}
\usepackage{etoolbox}
\usepackage{svg}
\usepackage[percent]{overpic}
\usepackage[round]{natbib}
\usepackage[colorinlistoftodos, shadow,color=blue!30!white
					,disable
]{todonotes}
\usepackage{xpatch}
\usepackage{siunitx}

\renewcommand{\sfdefault}{phv}

\fancypagestyle{first}{\fancyfoot[R]{\small\thepage}}

\setlist[itemize]{leftmargin=1em,itemsep=0ex,topsep=0ex}
\titlespacing*{\paragraph}{0pt}{0ex plus .1ex}{1ex}
\titlespacing*{\section}{0ex}{2.3ex plus .3ex minus .0ex}{.6ex plus .3ex minus .2ex}
\titlespacing*{\subsection}{0ex}{1.5ex plus .3ex minus .5ex}{.4ex plus .2ex minus .1ex}
\titlespacing*{\subsubsection}{0ex}{1.2ex plus .3ex minus .3ex}{.3ex plus .2ex minus .2ex}

\xapptocmd\normalsize{%
\abovedisplayskip=.8em plus .2em minus .2em
\belowdisplayskip=.6em plus .1em minus .1em
\abovedisplayshortskip=.8em plus .2em minus .2em
\belowdisplayshortskip=.6em plus .1em minus .1em
}{}{}

\setcitestyle{numbers}
\renewcommand{\cite}[1]{\citep{#1}}

\definecolor{mydarkblue}{rgb}{0.0,0.15,0.7}
\hypersetup{%
colorlinks=true,
linkcolor=mydarkblue,
citecolor=mydarkblue,
filecolor=mydarkblue,
urlcolor=mydarkblue}

\makeatletter

  \renewcommand{\maketitle}{%
    \begingroup
      {\centering\LARGE\@title\par}%
      \vskip 1em
      \centering
      \begin{tabular}[t]{@{}c@{}}\strut\@author\strut\end{tabular}%
      \vskip 0.3in minus 0.1in
    \endgroup
  }
\makeatother

\usepackage{xcolor}

\newcommand{\Mod}[1]{\ (\mathrm{mod}\ #1)}

\title{Hierarchical Reasoning Model}

\date{}
\author{%
\footnotesize
  Guan Wang$^{1,\dagger}$, %
  Jin Li$^{1,2}$, %
  Yuhao Sun$^{1}$, %
  Xing Chen$^{1}$, %
  Changling Liu$^{1}$, \\ %
\footnotesize
  Yue Wu$^{2}$, %
  Meng Lu$^{1,\dagger}$, %
  Sen Song$^{2,\dagger}$, %
  Yasin Abbasi Yadkori$^{1,\dagger}$%
  \\[1ex]
  $^{1}$Sapient Intelligence, Singapore
}

\begin{document}
\pagestyle{fancy}

\maketitle
\thispagestyle{first}

\footnotetext[2]{Tsinghua University $^{\dagger}$ Corresponding author. Contact: \texttt{research@sapient.inc}.

\hspace{\footnotesep}Code available at: \href{https://github.com/sapientinc/HRM}{\texttt{github.com/sapientinc/HRM}}}

\begin{abstract}
Reasoning, the process of devising and executing complex goal-oriented action sequences, remains a critical challenge in AI.
Current large language models (LLMs) primarily employ Chain-of-Thought (CoT) techniques, which suffer from brittle task decomposition, extensive data requirements, and high latency. Inspired by the hierarchical and multi-timescale processing in the human brain, we propose the Hierarchical Reasoning Model (HRM), a novel recurrent architecture that attains significant computational depth while maintaining both training stability and efficiency.
HRM executes sequential reasoning tasks in a single forward pass without explicit supervision of the intermediate process, through two interdependent recurrent modules: a high-level module responsible for slow, abstract planning, and a low-level module handling rapid, detailed computations. With only 27 million parameters, HRM achieves exceptional performance on complex reasoning tasks using only 1000 training samples. The model operates without pre-training or CoT data, yet achieves nearly perfect performance on challenging tasks including complex Sudoku puzzles and optimal path finding in large mazes.
Furthermore, HRM outperforms much larger models with significantly longer context windows on the Abstraction and Reasoning Corpus (ARC), a key benchmark for measuring artificial general intelligence capabilities.
These results underscore HRM’s potential as a transformative advancement toward universal computation and general-purpose reasoning systems.
\end{abstract}

\vfill
\begin{figure}[h]
  \centering
    \begin{tikzpicture}[inner sep=0pt, outer sep=0pt]
      \node[anchor=west] (A) at (0,0)
        {\includegraphics[width=0.35\linewidth]{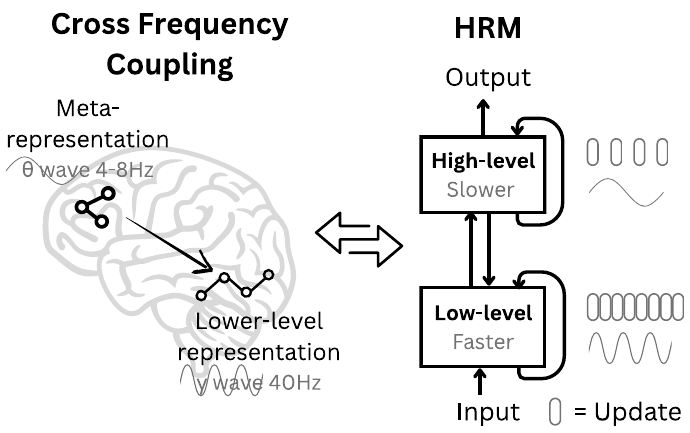}};
      \node[anchor=east, xshift=0.1in] (B) at (\linewidth,0)
        {\includegraphics[width=0.65\linewidth]{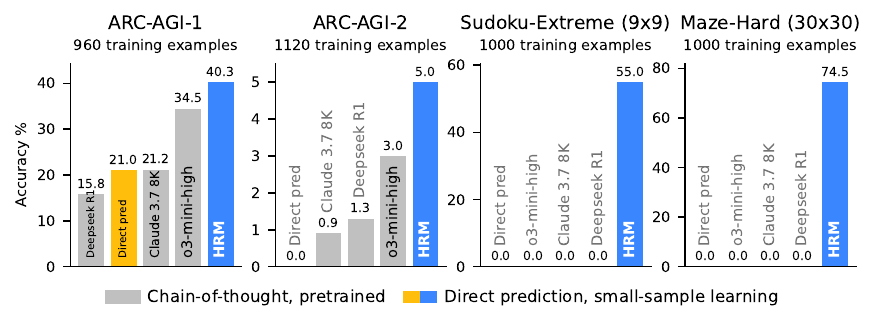}};
    \end{tikzpicture}
  \vspace*{-0.5ex}%
  \caption{\textbf{Left:} HRM is inspired by hierarchical processing and temporal separation in the brain. It has two recurrent networks operating at different timescales to collaboratively solve tasks. \textbf{Right:} With only about 1000 training examples, the HRM (\textasciitilde27M parameters) surpasses state-of-the-art CoT models on inductive benchmarks (ARC-AGI) and challenging symbolic tree-search puzzles (\textit{Sudoku-Extreme}, \textit{Maze-Hard}) where CoT models failed completely. The HRM was randomly initialized, and it solved the tasks directly from inputs without chain of thoughts.}
  \label{fig:benchmark_bars}
  \vspace*{-1ex}%
\end{figure}
\vfill
\clearpage

\section{Introduction}

\begin{figure}[t]
\centering
\includegraphics[width=0.9\linewidth]{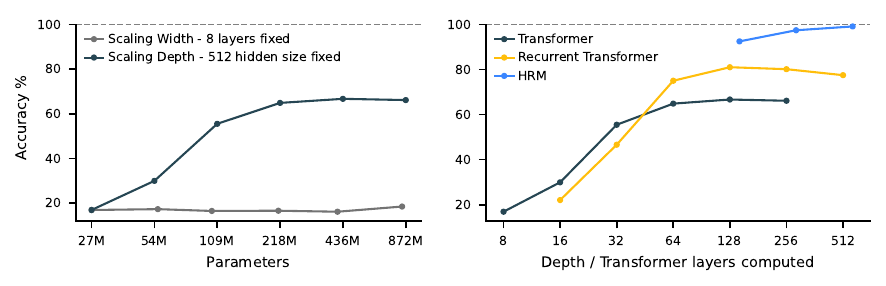} 
\caption{\textbf{The necessity of depth for complex reasoning.} \textbf{Left:} On \textit{Sudoku-Extreme Full}, which require extensive tree-search and backtracking, increasing a Transformer's width yields no performance gain, while increasing depth is critical. \textbf{Right:} Standard architectures saturates, failing to benefit from increased depth. HRM overcomes this fundamental limitation, effectively using its computational depth to achieve near-perfect accuracy.}
\label{fig:perf-layers}
\end{figure}

Deep learning, as its name suggests, emerged from the idea of stacking more layers to achieve increased representation power and improved performance~\cite{Goodfellow-et-al-2016, He2015DeepRL}. However, despite the remarkable success of large language models, their core architecture is paradoxically shallow~\cite{strobl-2023}. This imposes a fundamental constraint on their most sought-after capability: reasoning. The fixed depth of standard Transformers places them in computational complexity classes such as $AC^0$ or $TC^0$~\cite{10.5555/1631171.1631212}, preventing them from solving problems that require polynomial time~\cite{MS-2023, Chiang-2025}. LLMs are not Turing-complete and thus they cannot, at least in a purely end-to-end manner, execute complex algorithmic reasoning that is necessary for deliberate planning or symbolic manipulation tasks~\cite{Lehnert2024BeyondAB, Bounsi2024TransformersMN}. For example, our results on the Sudoku task show that increasing Transformer model depth {\it can} improve performance,\footnote{Simply increasing the model width does not improve performance here.} but performance remains far from optimal even with very deep models (see \Cref{fig:perf-layers}), which supports the conjectured limitations of the LLM scaling paradigm~\citep{merrill-sabharwal-2023-parallelism}.

The LLMs literature has relied largely on Chain-of-Thought (CoT) prompting for reasoning~\cite{ChainOfThought2022}. CoT externalizes reasoning into token-level language by breaking down complex tasks into simpler intermediate steps, sequentially generating text using a shallow model~\cite{MS-2024}. However, CoT for reasoning is a crutch, not a satisfactory solution. It relies on brittle, human-defined decompositions where a single misstep or a misorder of the steps can derail the reasoning process entirely~\cite{Chen2024PremiseOM, Xu2024PreemptiveA}. This dependency on explicit linguistic steps tethers reasoning to 
patterns at the token level. As a result, CoT reasoning often requires significant amount of training data and generates a large number of tokens for complex reasoning tasks, resulting in slow response times. A more efficient approach is needed to minimize these data requirements~\cite{Villalobos2022WillWR}.

Towards this goal, we explore ``latent reasoning'', where the model conducts computations within its internal hidden state space~\cite{Chen2025ReasoningBL, CoconutLatentReasoning2024}. This aligns with the understanding that language is a tool for human communication, not the substrate of thought itself~\cite{fedorenko2024language}; the brain sustains lengthy, coherent chains of reasoning with remarkable efficiency in a latent space, without constant translation back to language. However, the power of latent reasoning is still fundamentally constrained by a model's \textit{effective computational depth}. Naively stacking layers is notoriously difficult due to vanishing gradients, which plague training stability and effectiveness~\citep{Goodfellow-et-al-2016, wang2024deepnet}. Recurrent architectures, a natural alternative for sequential tasks, often suffer from early convergence, rendering subsequent computational steps inert, and rely on the biologically implausible, computationally expensive and memory intensive Backpropagation Through Time (BPTT) for training~\cite{LILLICRAP201982}.

The human brain provides a compelling blueprint for achieving the effective computational depth that contemporary artificial models lack. It organizes computation hierarchically across cortical regions operating at different timescales, enabling deep, multi-stage reasoning~\citep{murray2014hierarchy, zeraati2023intrinsic, huntenburg2018large}. Recurrent feedback loops iteratively refine internal representations, allowing slow, higher-level areas to guide, and fast, lower-level circuits to execute—subordinate processing while preserving global coherence~\citep{lamme2000distinct, bastos2012canonical, kaleb2024feedback}. Notably, the brain achieves such depth without incurring the prohibitive credit-assignment costs that typically hamper recurrent networks from backpropagation through time~\citep{LILLICRAP201982, lillicrap2020backpropagation}.

Inspired by this hierarchical and multi-timescale biological architecture, we propose the Hierarchical Reasoning Model (HRM). HRM is designed to significantly increase the effective computational depth. It features two coupled recurrent modules: a high-level (H) module for abstract, deliberate reasoning, and a low-level (L) module for fast, detailed computations. This structure avoids the rapid convergence of standard recurrent models through a process we term ``hierarchical convergence.'' The slow-updating H-module advances only after the fast-updating L-module has completed multiple computational steps and reached a local equilibrium, at which point the L-module is reset to begin a new computational phase.

Furthermore, we propose a one-step gradient approximation for training HRM, which offers improved efficiency and eliminates the requirement for BPTT. This design maintains a constant memory footprint ($O(1)$ compared to BPTT's $O(T)$ for $T$ timesteps) throughout the backpropagation process, making it scalable and more biologically plausible.

Leveraging its enhanced effective depth, HRM excels at tasks that demand extensive search and backtracking. \textbf{Using only 1,000 input-output examples, without pre-training or CoT supervision}, HRM learns to solve problems that are intractable for even the most advanced LLMs. For example, it achieves near-perfect accuracy in complex Sudoku puzzles (\textit{Sudoku-Extreme Full}) and optimal pathfinding in 30x30 mazes, where state-of-the-art CoT methods completely fail (0\% accuracy). In the Abstraction and Reasoning Corpus (ARC) AGI Challenge~\cite{AbstractionReasoning2019, Chollet2024ARCP2, Chollet2025ARCAGI2AN} - a benchmark of inductive reasoning - HRM, trained from scratch with only the official dataset (\textasciitilde1000 examples), with only 27M parameters and a 30x30 grid context (900 tokens), achieves a performance of \textbf{40.3\%}, which substantially surpasses leading CoT-based models like o3-mini-high (34.5\%) and Claude 3.7 8K context (21.2\%), despite their considerably larger parameter sizes and context lengths, as shown in \Cref{fig:benchmark_bars}. This represents a promising direction toward the development of next-generation AI reasoning systems with universal computational capabilities.

\section{Hierarchical Reasoning Model}

We present the HRM, inspired by three fundamental principles of neural computation observed in the brain:

\begin{itemize}
  \item \textbf{Hierarchical processing:} The brain processes information across a hierarchy of cortical areas. Higher-level areas integrate information over longer timescales and form abstract representations, while lower-level areas handle more immediate, detailed sensory and motor processing~\citep{murray2014hierarchy, huntenburg2018large,zeraati2023intrinsic}.
  \item \textbf{Temporal Separation:} These hierarchical levels in the brain operate at distinct intrinsic timescales, reflected in neural rhythms (e.g., slow theta waves, 4–8 Hz and fast gamma waves, 30–100 Hz)~\citep{buzsaki2000gamma, buzsaki2006rhythms}. This separation allows for stable, high-level guidance of rapid, low-level computations~\citep{pahor2014theta, tort2009theta}.
  \item \textbf{Recurrent Connectivity:} The brain features extensive recurrent connections. These feedback loops enable iterative refinement, yielding more accurate and context-sensitive representations at the cost of additional processing time. Additionally, the brain largely avoids the problematic deep credit assignment problem associated with BPTT~\citep{LILLICRAP201982}.
\end{itemize}

The HRM model consists of four learnable components: an input network $f_I(\cdot; \theta_I)$, a low-level recurrent module $f_L(\cdot; \theta_L)$, a high-level recurrent module $f_H(\cdot; \theta_H)$, and an output network $f_O(\cdot; \theta_O)$. The model's dynamics unfold over $N$ high-level cycles of $T$ low-level timesteps each\footnote{While inspired by temporal separation in the brain, our model's ``high-level'' and ``low-level'' modules are conceptual abstractions and do not map directly to specific neural oscillation frequencies.}. We index the total timesteps of one forward pass by $i = 1, \dots, N \times T$. The modules $f_L$ and $f_H$ each keep a hidden state---$z_L^i$ for $f_L$ and $z_H^i$ for $f_H$---which are initialized with the vectors $z_L^0$ and $z_H^0$, respectively.  

The HRM maps an input vector $x$ to an output prediction vector $\hat{y}$ as follows. First, the input $x$ is projected into a working representation $\tilde{x}$ by the input network:
\begin{equation*}
    \tilde{x} = f_I(x; \theta_I) \;.
\end{equation*}
At each timestep $i$, the L-module updates its state conditioned on its own previous state, the H-module’s current state (which remains fixed throughout the cycle), and the input representation. The H-module only updates once per cycle (i.e., every $T$ timesteps) using the L-module’s final state at the end of that cycle:
\begin{align*}
    z_L^i &= f_L\left( z_L^{i-1}, z_H^{i-1}, \tilde{x}; \theta_L \right)\,, \\
    z_H^i &= \begin{cases}
f_H\left( z_H^{i-1}, z_L^{i}; \theta_H \right) & \text{if } i \equiv 0\Mod{T}\,, \\
z_H^{i-1} & \text{otherwise }\;.
\end{cases}
\end{align*}

Finally, after $N$ full cycles, a prediction $\hat{y}$ is extracted from the hidden state of the H-module:

\begin{equation*}
    \hat{y} = f_O(z_H^{N T}; \theta_O) \;.
\end{equation*}

This entire $NT$-timestep process represents a single forward pass of the HRM. A halting mechanism (detailed later in this section) determines whether the model should terminate, in which case $\hat{y}$ will be used as the final prediction, or continue with an additional forward pass. 

\paragraph{Hierarchical convergence}
\begin{figure}[t]
\centering
\includegraphics[width=\linewidth]{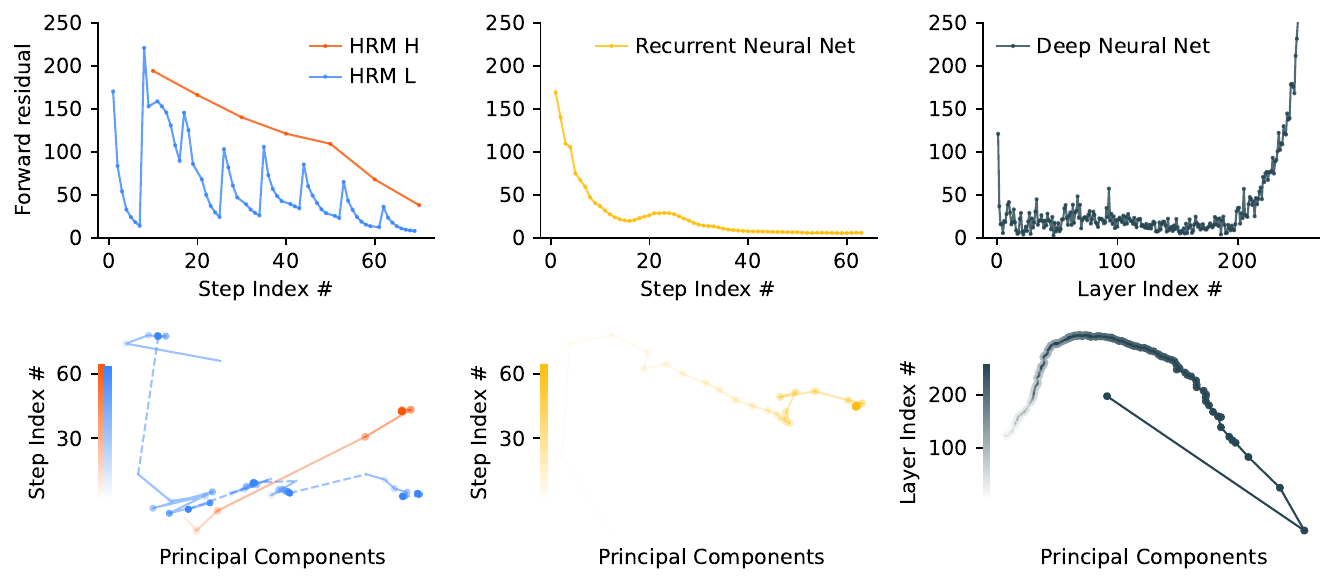}
\caption{Comparison of forward residuals and PCA trajectories. HRM shows hierarchical convergence: the H-module steadily converges, while the L-module repeatedly converges within cycles before being reset by H, resulting in residual spikes. The recurrent neural network exhibits rapid convergence with residuals quickly approaching zero. In contrast, the deep neural network experiences vanishing gradients, with significant residuals primarily in the initial (input) and final layers.}
\label{fig:forward-residual}
\end{figure}

Although convergence is crucial for recurrent networks, standard RNNs are fundamentally limited by their tendency to converge too early. As the hidden state settles toward a fixed point, update magnitudes shrink, effectively stalling subsequent computation and capping the network’s effective depth. To preserve computational power, we actually want convergence to proceed very slowly--but engineering that gradual approach is difficult, since pushing convergence too far edges the system toward instability.

HRM is explicitly designed to counteract this premature convergence through a process we term \textit{hierarchical convergence}. During each cycle, the L-module (an RNN) exhibits stable convergence to a \textit{local equilibrium}. This equilibrium, however, depends on the high-level state $z_H$ supplied during that cycle. After completing the $T$ steps, the H-module incorporates the sub-computation's outcome (the final state $z_L$) and performs its own update. This $z_H$ update establishes a fresh context for the L-module, essentially ``restarting'' its computational path and initiating a new convergence phase toward a different local equilibrium.

This process allows the HRM to perform a sequence of distinct, stable, nested computations, where the H-module directs the overall problem-solving strategy and the L-module executes the intensive search or refinement required for each step. Although a standard RNN may approach convergence within $T$ iterations, the hierarchical convergence benefits from an enhanced effective depth of $NT$ steps. As empirically shown in \Cref{fig:forward-residual}, this mechanism allows HRM both to maintain high computational activity (forward residual) over many steps (in contrast to a standard RNN, whose activity rapidly decays) and to enjoy stable convergence. This translates into better performance at any computation depth, as illustrated in \Cref{fig:perf-layers}.

\paragraph{Approximate gradient}
Recurrent models typically use BPTT to compute gradients. However, BPTT requires storing the hidden states from the forward pass and then combining them with gradients during the backward pass, which demands $O(T)$ memory for T timesteps. This heavy memory burden forces smaller batch sizes and leads to poor GPU utilization, especially for large-scale networks. Additionally, because retaining the full history trace through time is biologically implausible, it is unlikely that the brain implements BPTT~\cite{LILLICRAP201982}. 

Fortunately, if a recurrent neural network converges to a fixed point, we can avoid unrolling its state sequence by applying backpropagation in a single step at that equilibrium point. Moreover, such a mechanism could plausibly be implemented in the brain using only local learning rules~\cite{Scellier2016EquilibriumPB, Eprop}. Based on this finding, we propose a one-step approximation of the HRM gradient--using the gradient of the last state of each module and treating other states as constant. The gradient path is, therefore, 
{\small\begin{quote}
    Output head → final state of the H-module → final state of the L-module → input embedding
\end{quote}}

\begin{wrapfigure}{r}{0.4\textwidth}
  \vspace{-0.3in}
  \centering
  \begin{minipage}[t]{\linewidth}
    \includegraphics[width=\linewidth]{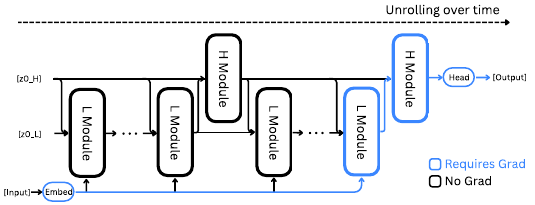}
    \begin{lstlisting}[language=python]
def hrm(z, x, N=2, T=2):
    x = input_embedding(x)
    zH, zL = z

    with torch.no_grad():
        for _i in range(N * T - 1):
            zL = L_net(zL, zH, x)
            if (_i + 1) % T == 0:
                zH = H_net(zH, zL)

    # 1-step grad
    zL = L_net(zL, zH, x)
    zH = H_net(zH, zL)
    return (zH, zL), output_head(zH)

# Deep Supervision
for x, y_true in train_dataloader:
    z = z_init
    for step in range(N_supervision):
        z, y_hat = hrm(z, x)
        
        loss = softmax_cross_entropy(y_hat, y_true)
        z = z.detach()

        loss.backward()
        opt.step()
        opt.zero_grad()
    \end{lstlisting}
  \end{minipage}
  \vspace{-0.15in}
  \caption{\textbf{Top:} Diagram of HRM with approximate gradient. \textbf{Bottom:} Pseudocode of HRM with deep supervision training in PyTorch.}
  \label{fig:pseudocode}
\end{wrapfigure}

The above method needs $O(1)$ memory, does not require unrolling through time, and can be easily implemented with an autograd framework such as PyTorch, as shown in \Cref{fig:pseudocode}. Given that each module only needs to back-propagate errors through its most recent local synaptic activity, this approach aligns well with the perspective that cortical credit assignment relies on short-range, temporally local mechanisms rather than on a global replay of activity patterns.

The one-step gradient approximation is theoretically grounded in the mathematics of Deep Equilibrium Models (DEQ)~\citep{bai2019deep} which employs the Implicit Function Theorem (IFT) to bypass BPTT, as detailed next. Consider an idealized HRM behavior where, during high-level cycle $k$, the L-module repeatedly updates until its state $z_L$ converges to a local fixed point $z_L^\star$. This fixed point, given the current high-level state $z_H^{k-1}$, can be expressed as
\begin{equation*}
    z_L^\star = f_L(z_L^\star, z_H^{k-1}, \tilde{x}; \theta_L) \;.
\end{equation*}
The H-module then performs a single update using this converged L-state:
\begin{equation*}
    z_H^k = f_H(z_H^{k-1}, z_L^\star; \theta_H) \;.
\end{equation*}
With a proper mapping $\mathcal{F}$, the updates to the high-level state can be written in a more compact form as $z_H^k = \mathcal{F}(z_H^{k-1}; \tilde{x}, \theta)$, where $\theta = (\theta_H, \theta_L)$, and the fixed-point can be written as $z_H^\star = \mathcal{F}(z_H^\star; \tilde{x}, \theta)$. Let $ J_{\mathcal{F}} = \frac{\partial \mathcal{F}}{\partial z_H}$ be the Jacobian of $\mathcal{F}$, and assume that the matrix $I - J_{\mathcal{F}}$ is invertible at $z_H^\star$ and that the mapping $\mathcal{F}$ is continuously differentiable. The Implicit Function Theorem then allows us to calculate the exact gradient of fixed point $z_H^\star$ with respect to the parameters $\theta$ without explicit backpropagation:
\begin{equation}
\label{eq:IFT-zH}
\frac{\partial z_H^\star}{\partial \theta} = \left( I - J_{\mathcal{F}} \big|_{z_H^\star} \right)^{-1} \frac{\partial \mathcal{F}}{\partial \theta} \bigg|_{z_H^\star} \;.
\end{equation}

Calculating the above gradient requires evaluating and inverting matrix $(I - J_{\mathcal{F}})$ that can be computationally expensive. Given the Neumann series expansion,
\begin{equation*}
(I - J_{\mathcal{F}})^{-1} = I + J_{\mathcal{F}} + J_{\mathcal{F}}^2 + J_{\mathcal{F}}^3 + \dots \,,
\end{equation*}
the so-called \textit{1-step gradient}~\cite{Geng2021OnTI} approximates the series by considering only its first term, i.e. $(I - J_{\mathcal{F}})^{-1} \approx I$, and leads to the following approximation of \Cref{eq:IFT-zH}:  
\begin{equation}
\label{eq:1-step-1}
\frac{\partial z_H^*}{\partial\theta_H} \approx \frac{\partial f_H}{\partial \theta_H},
\quad
\frac{\partial z_H^*}{\partial\theta_L}\approx \frac{\partial f_H}{\partial z^*_L} \cdot \frac{\partial z^*_L}{\partial \theta_L},
\quad
\frac{\partial z_H^*}{\partial\theta_I} \approx \frac{\partial f_H}{\partial z^*_L} \cdot
\frac{\partial z^*_L}{\partial \theta_I} \;.
\end{equation}
The gradients of the low-level fixed point, $\frac{\partial z^*_L}{\partial \theta_L}$ and $\frac{\partial z^*_L}{\partial \theta_I}$, can also be approximated using another application of the 1-step gradient:
\begin{equation}
\label{eq:1-step-2}
\frac{\partial z^*_L}{\partial \theta_L} \approx \frac{\partial f_L}{\partial \theta_L},
\quad
\frac{\partial z^*_L}{\partial \theta_I} \approx \frac{\partial f_L}{\partial \theta_I} \;.
\end{equation}
By substituting \Cref{eq:1-step-2} back into \Cref{eq:1-step-1}, we arrive at the final simplified gradients.

Before defining our loss function, we must first introduce two key elements of our proposed method: \textit{deep supervision} and \textit{adaptive computational time}.

\paragraph{Deep supervision}
Inspired by the principle that periodic neural oscillations regulate when learning occurs in the brain~\cite{BEGUS2020100810}, we incorporate a deep supervision mechanism into HRM, as detailed next. 

Given a data sample $(x, y)$, we run multiple forward passes of the HRM model, each of which we refer to as a \textit{segment}. Let $M$ denote the total number of segments executed before termination. For each segment $m \in \{1, \dots, M\}$, let $z^m = (z^{mNT}_H, z^{mNT}_L)$ represent the hidden state at the conclusion of segment $m$, encompassing both high-level and low-level state components.

At each segment $m$, we apply a deep supervision step as follows:
\begin{enumerate}
\vspace{-0.15in}
    \item Given the state $z^{m-1}$ from the previous segment, compute the next state $z^m$ and its associated output $\hat{y}^m$ through a forward pass in the HRM model:
    \begin{equation*}
    (z^m, \hat{y}^m) \leftarrow \textsc{HRM}(z^{m-1}, x; \theta)
    \end{equation*}
    \item Compute the loss for the current segment:
    \begin{equation*}
    L^m \leftarrow \textsc{Loss}(\hat{y}^m, y)
    \end{equation*}
    \item Update parameters:
    \begin{equation*}
    \theta \leftarrow \textsc{OptimizerStep}(\theta, \nabla_{\theta} L^m)
    \end{equation*}
\end{enumerate}
The crucial aspect of this procedure is that the hidden state $z^m$ is ``detached'' from the computation graph before being used as the input state for the next segment. Consequently, gradients from segment $m+1$ do not propagate back through segment $m$, effectively creating a 1-step approximation of the gradient of the recursive deep supervision process~\citep{DEQ-Flow, Ramzi2021SHINEST}. This approach provides more frequent feedback to the H-module and serves as a regularization mechanism, demonstrating superior empirical performance and enhanced stability in deep equilibrium models when compared to more complex, Jacobian-based regularization techniques~\citep{DEQ-Flow, Bai2021StabilizingEM}. \Cref{fig:pseudocode} shows pseudocode of deep supervision training.

\paragraph{Adaptive computational time (ACT)}
The brain dynamically alternates between automatic thinking (``System 1'') and deliberate reasoning (``System 2'')~\citep{kahneman2011thinking}. Neuroscientific evidence shows that these cognitive modes share overlapping neural circuits, particularly within regions such as the prefrontal cortex and the default mode network~\citep{lieberman2007social,buckner2008brain}. This indicates that the brain dynamically modulates the ``runtime'' of these circuits according to task complexity and potential rewards~\citep{raichle2015brain,westbrook2015cognitive}.

Inspired by the above mechanism, we incorporate an adaptive halting strategy into HRM that enables ``thinking, fast and slow''. This integration leverages deep supervision and uses the Q-learning algorithm~\cite{Sutton-Barto-2018} to adaptively determine the number of segments. A Q-head uses the final state of the H-module to predict the Q-values $\hat{Q}^m = (\hat{Q}^m_\text{halt}, \hat{Q}^m_\text{continue})$ of the ``halt'' and ``continue'' actions:
\begin{equation*}
    \hat{Q}^m = \sigma(\theta_Q^\top z_H^{mNT})\,,
\end{equation*}
where $\sigma$ denotes the sigmoid function applied element-wise. The halt or continue action is chosen using a randomized strategy as detailed next. Let $M_{\max}$ denote the maximum number of segments (a fixed hyperparameter) and $M_{\min}$ denote the minimum number of segments (a random variable). The value of $M_{\min}$ is determined stochastically: with probability $\varepsilon$, it is sampled uniformly from the set $\{2,\cdots,M_{\max}\}$ (to encourage longer thinking), and with probability $1-\varepsilon$, it is set to 1. The halt action is selected under two conditions: when the segment count surpasses the maximum threshold $M_{\max}$, or when the estimated halt value $\hat{Q}_\text{halt}$ exceeds the estimated continue value $\hat{Q}_\text{continue}$ and the segment count has reached at least the minimum threshold $M_{\min}$.

The Q-head is updated through a Q-learning algorithm, which is defined on the following episodic Markov Decision Process (MDP). The state of the MDP at segment $m$ is $z^m$, and the action space is $\{\text{halt}, \text{continue}\}$. Choosing the action ``halt'' terminates the episode and returns a binary reward indicating prediction correctness, i.e., $\mathbf{1}\{\hat{y}^m = y\}$. Choosing ``continue'' yields a reward of 0 and the state transitions to $z^{m+1}$. Thus, the Q-learning targets for the two actions $\hat{G}^m = (\hat{G}^m_\text{halt}, \hat{G}^m_\text{continue})$ are given by
\begin{align*}
\hat{G}_\text{halt}^m &= \mathbf{1}\{\hat{y}^m = y\}\,, \\
\hat{G}_\text{continue}^m &= 
\begin{cases}
\hat{Q}_\text{halt}^{m+1}, & \text{if $m \geq M_{\max}$}\,,\\[6pt]
\max(\hat{Q}_\text{halt}^{m+1}, \hat{Q}_\text{continue}^{m+1})\,, & \text{otherwise}\;.
\end{cases}
\end{align*}
We can now define the loss function of our learning procedure. The overall loss for each supervision segment combines both the Q-head loss and the sequence-to-sequence loss:
\begin{equation*}
    L^m_\text{ACT} = \textsc{Loss}(\hat{y}^m, y) + \textsc{BinaryCrossEntropy}(\hat{Q}^m, \hat{G}^m) \;.
\end{equation*}
Minimizing the above loss enables both accurate predictions and nearly optimal stopping decisions. 

Selecting the ``halt'' action ends the supervision loop. In practice, sequences are processed in batches, which can be easily handled by substituting any halted sample in the batch with a fresh sample from the dataloader.

\begin{figure}[t!]
\centering
\includegraphics[width=\linewidth]{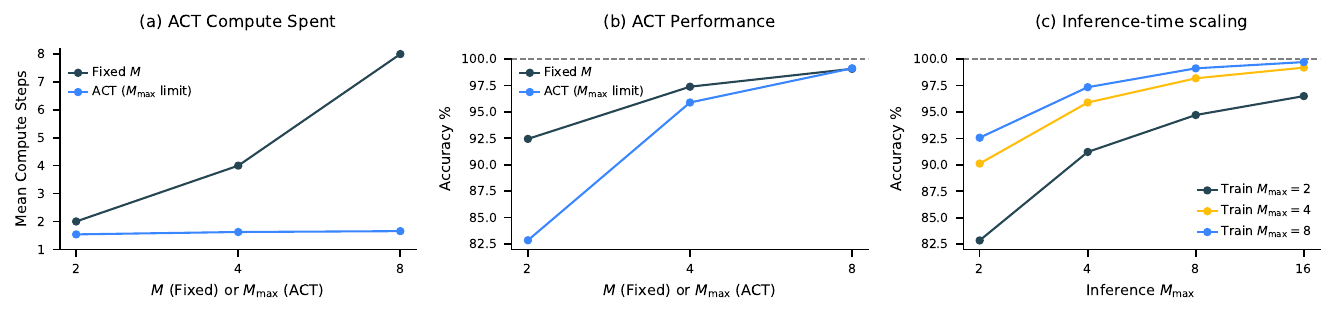} 
\caption{\textbf{Effectiveness of Adaptive Computation Time (ACT)} on the \textit{Sudoku-Extreme-Full}.
\textbf{(a)} Mean compute steps used by models with ACT versus models with a fixed number of compute steps ($M$). ACT maintains a low and stable number of average compute steps even as the maximum limit ($M_{\max}$) increases.
\textbf{(b)} Accuracy comparison. The ACT model achieves performance comparable to the fixed-compute model while utilizing substantially fewer computational steps on average.
\textbf{(c)} Inference-time scalability. Models trained with a specific $M_{\max}$ can generalize to higher computational limits during inference, leading to improved accuracy. For example, a model trained with $M_{\max}=8$ continues to see accuracy gains when run with $M_{\max}=16$ during inference.}
\label{fig:act}
\end{figure}

\Cref{fig:act} presents a performance comparison between two HRM variants: one incorporating ACT and another employing a fixed computational step count equivalent to ACT's $M_{\max}$ parameter. It shows that ACT effectively adapts its computational resources based on task complexity, achieving significant computational savings with minimal impact on performance.

\paragraph{Inference-time scaling}
An effective neural model should exploit additional computational resources during inference to enhance performance. As illustrated in \Cref{fig:act}-(c), HRM seamlessly achieves inference-time scaling by simply increasing the computational limit parameter, $M_{\max}$ without requiring further training or architectural modifications.

Additional compute is especially effective for tasks that demand deeper reasoning. On Sudoku---a problem that often requires long-term planning---HRM exhibits strong inference-time scaling. On the other hand, we find that extra computational resources yield minimal gains in ARC-AGI challenge, as solutions generally require only a few transformations.

\paragraph{Stability of Q-learning in ACT}
The deep Q-learning that underpins our ACT mechanism is known to be prone to instability, often requiring stabilization techniques such as replay buffers and target networks~\citep{DQN}, which are absent in our design. Our approach, however, achieves stability through the intrinsic properties of our model and training procedure. Recent theoretical work by \citet{PQN2025} shows that Q-learning can achieve convergence if network parameters are bounded, weight decay is incorporated during training, and post-normalization layers are implemented.
Our model satisfies these conditions through its Post-Norm architecture that employs RMSNorm (a layer normalization variant) and the AdamW optimizer. AdamW has been shown to solve an $L_\infty$-constrained optimization problem, ensuring that model parameters remain bounded by $1/\lambda$ \citep{Xie2024ImplicitBO}.

\paragraph{Architectural details}
We employ a sequence-to-sequence architecture for HRM. Both input and output are represented as token sequences: $x = (x_1,\ldots,x_l)$ and $y = (y_1,\ldots,y_{l^\prime})$ respectively. The model includes an embedding layer $f_I$ that converts discrete tokens into vector representations, and an output head $f_O(z; \theta_O) = \text{softmax}(\theta_O z)$ that transforms hidden states into token probability distributions $\hat{y}$. For small-sample experiments, we replace softmax with stablemax~\cite{PBMB-2025} to improve generalization performance. The sequence-to-sequence loss is averaged over all tokens, $\textsc{Loss}(\hat{y}, y) = \frac{1}{l^\prime} \sum_{i=1}^{l^\prime}\log p(y_i)$, where $p(y_i)$ is the probability that distribution $\hat{y}_i$ assigns to token $y_i$. The initial hidden states $z^0$ are initialized by sampling from a truncated normal distribution with standard deviation of 1, truncation of 2, and kept fixed throughout training.

Both the low-level and high-level recurrent modules $f_L$ and $f_H$ are implemented using encoder-only Transformer~\cite{vaswani2017attention} blocks with identical architectures and dimensions. 
These modules take multiple inputs, and we use straightforward element-wise addition to combine them, though more sophisticated merging techniques such as gating mechanisms could potentially improve performance and is left for future work. For all Transformer blocks in this work---including those in the baseline models---we incorporate the enhancements found in modern LLMs (based on Llama~\cite{meta2024llama3} architectures). These improvements include Rotary Positional Encoding~\citep{SHLPBL-2024}, Gated Linear Units~\cite{Shazeer2020GLUVI}, RMSNorm~\cite{Zhang2019RootMS}, and the removal of bias terms from linear layers. 

Furthermore, both HRM and recurrent Transformer models implement a Post-Norm architecture with weights initialized via truncated LeCun Normal initialization~\cite{Klambauer2017SelfNormalizingNN, jax_lecun_normal_initializer,lecun2002efficient}, while the scale and bias parameters are excluded from RMSNorm. All parameters are optimized using the Adam-atan2 optimizer~\citep{EXWANLGSKLP-2024}, a scale-invariant variant of Adam~\citep{KB-2017}, combined with a constant learning rate that includes linear warm-up.
\section{Results}

This section begins by describing the ARC-AGI, Sudoku, and Maze benchmarks, followed by an overview of the baseline models and their results. \Cref{fig:benchmark_intro}-(a,b,c) presents a visual representation of the three benchmark tasks, which are selected to evaluate various reasoning abilities in AI models.

\subsection{Benchmarks}

\begin{figure}[t]
    \centering
    \includegraphics[width=1\linewidth]{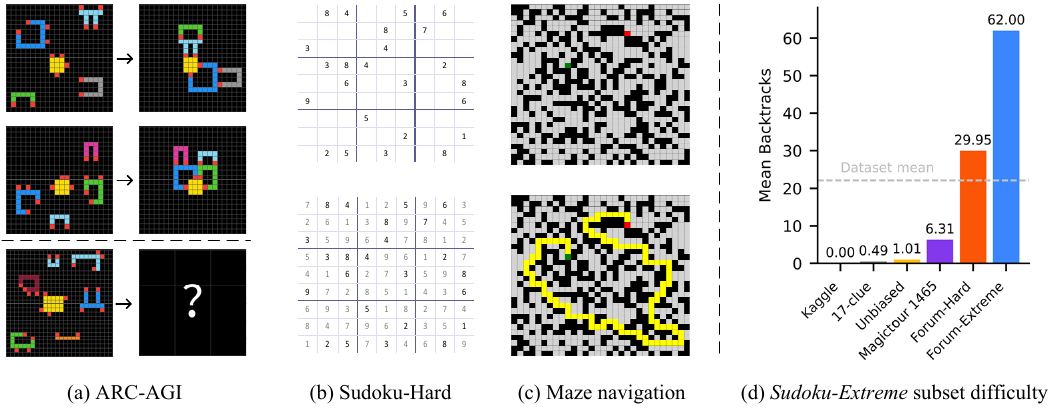}
    \caption{\textbf{Left:} Visualization of benchmark tasks. \textbf{Right:} Difficulty of \textit{Sudoku-Extreme} examples.}
    \label{fig:benchmark_intro}
\end{figure}


\textbf{ARC-AGI Challenge}~ The ARC-AGI benchmark evaluates general fluid intelligence through IQ-test-like puzzles that require inductive reasoning~\cite{AbstractionReasoning2019}. The initial version, ARC-AGI-1, presents challenges as input-label grid pairs that force AI systems to extract and generalize abstract rules from just a few examples. Each task provides a few input–output demonstration pairs (usually 2–3) and a test input. An AI model has two attempts to produce the correct output grid. Although some believe that mastering ARC-AGI would signal true artificial general intelligence, its primary purpose is to expose the current roadblocks in AGI progress. In fact, both conventional deep learning methods and CoT techniques have faced significant challenges with ARC-AGI-1, primarily because it requires the ability to generalize to entirely new tasks~\citep{Chollet2024ARCP2}.

Addressing the limitations identified in ARC-AGI-1, ARC-AGI-2 significantly expands the benchmark by providing a more comprehensive and carefully refined collection of tasks. These new tasks emphasize deeper compositional reasoning, multi-step logic, contextual rule application, and symbolic abstraction. Human calibration studies show these tasks are challenging but doable for people, while being much harder for current AI systems, offering a clearer measure of general reasoning abilities~\citep{Chollet2025ARCAGI2AN}.

\textbf{Sudoku-Extreme}~ Sudoku is a 9$\times$9 logic puzzle, requiring each row, column, and 3$\times$3 block to contain the digits 1–9 exactly once. A prediction is considered correct if it exactly matches the puzzle's unique solution. Sudoku's complex logical structure makes it a popular benchmark for evaluating logical reasoning in machine learning~\citep{Palm2017RecurrentRN,Long2023LargeLM,Du2024LearningIR}.

The most frequently used Sudoku dataset in research, namely the Kaggle dataset~\citep{sudoku2018}, can be fully solved using elementary single-digit techniques~\citep{SDT-Sudoku}. The minimal 17-clue puzzles~\citep{Palm2017RecurrentRN}, another widely-used collection, might seem more challenging due to its small number of clues. However, this perception is misleading---since 17 represents the minimum number of clues required to guarantee a unique Sudoku solution, these hints need to be highly orthogonal to each other. This orthogonal arrangement leads to many direct, easily-resolved solution paths~\cite{tdoku}.

We introduce \textit{Sudoku-Extreme}, a more challenging dataset that is compiled from the aforementioned easy datasets as well as puzzles recognized by the Sudoku community as exceptionally difficult for human players:
\begin{itemize}
    \item Easy puzzles compiled from Kaggle, 17-clue, plus unbiased samples from the Sudoku puzzle distribution~\cite{tdoku}: totaling \num{1149158} puzzles.
    \item Challenging puzzles compiled from Magictour 1465, Forum-Hard and Forum-Extreme subsets: totaling \num{3104157} puzzles.
\end{itemize}
The compiled data then undergo a strict 90/10 train-test split, ensuring that the test set puzzles cannot be derived through equivalent transformations of any training samples. \textit{Sudoku-Extreme} is a down-sampled subset of this data containing 1000 training examples. We use \textit{Sudoku-Extreme}  in our main experiments (\Cref{fig:benchmark_bars}), which focuses on small-sample learning scenarios. To guarantee convergence and control overfitting effects in our analysis experiments (\Cref{fig:perf-layers,fig:forward-residual,fig:act}), we use the complete training data, \textit{Sudoku-Extreme-Full}, containing \num{3831994} examples.


We measure puzzle difficulty by counting the number of search backtracks (``guesses'') required by a smart Sudoku solver program \textit{tdoku}, which uses propositional logic to reduce the number of guesses~\citep{tdoku}. Our \textit{Sudoku-Extreme} dataset exhibits a mean difficulty of $22$ backtracks per puzzle, significantly higher than existing datasets, including recent handmade puzzles Sudoku-Bench \citep{Seely2025SudokuBenchEC} which average just $0.45$ backtracks per puzzle. These subset complexity levels are shown in~\Cref{fig:benchmark_intro}-(d).

\textbf{Maze-Hard} This task involves finding the optimal path in a 30$\times$30 maze, making it interpretable and frequently used for training LLMs in search tasks \citep{Darlow2025ContinuousTM, dualformer2025, searchformer2024}. 
We adopt the instance generation procedure of \citet{searchformer2024}, but introduce an additional filter to retain only those instances whose difficulty exceeds 110. Here, ``difficulty'' is defined as the length of the shortest path, which aligns with the linear time complexity of the wavefront breadth-first search algorithm on GPUs~\citep{wavefrontBFS}. A path is considered correct if it is valid and optimal---that is, the shortest route from the start to the goal. The training and test set both include 1000 examples.

\subsection{Evaluation Details}

For all benchmarks, HRM models were initialized with random weights and trained in the sequence-to-sequence setup using the input-output pairs. The two-dimensional input and output grids were flattened and then padded to the maximum sequence length. The resulting performance is shown in \Cref{fig:benchmark_bars}. Remarkably, HRM attains these results with just \textasciitilde1000 training examples per task---and \textbf{without pretraining or CoT labels}.

For ARC-AGI challenge, we start with (1) all demonstration and test input-label pairs from the training set, and (2) all demonstration pairs along with test inputs from the evaluation set. The dataset is augmented by applying translations, rotations, flips, and color permutations to the puzzles. Each task example is prepended with a learnable special token that represents the puzzle it belongs to. At test time, we proceed as follows for each test input in the evaluation set: (1) Generate and solve 1000 augmented variants and, for each, apply the inverse‐augmentation transform to obtain a prediction. (2) Choose the two most popular predictions as the final outputs.\footnote{The ARC-AGI allows two attempts for each test input.} All reported results are obtained by comparing the outputs with the withheld test labels from the evaluation set.


We augment Sudoku puzzles by applying band and digit permutations, while data augmentation is disabled for Maze tasks. Both tasks undergo only a single inference pass.

For ARC-AGI, the scores of the CoT models are taken from the official leaderboard~\citep{Chollet2025ARCAGI2AN}, while for Sudoku and Maze, the scores are obtained by evaluating through the corresponding API.


In \Cref{fig:benchmark_bars}, the baselines are grouped based on whether they are pre-trained and use CoT, or neither. The ``Direct pred'' baseline means using ``direct prediction without CoT and pre-training'', which retains the exact training setup of HRM but swaps in a Transformer architecture. Interestingly, on ARC-AGI-1, ``Direct pred'' matches the performance of \citet{liao2025arcagiwithoutpretraining}, who built a carefully designed, domain-specific equivariant network for learning the ARC-AGI task from scratch, without pre-training. By substituting the Transformer architecture with HRM's hierarchical framework and implementing ACT, we achieve more than a twofold performance improvement.

On the \textit{Sudoku-Extreme} and \textit{Maze-Hard} benchmarks, the performance gap between HRM and the baseline methods is significant, as the baselines almost never manage to solve the tasks. These benchmarks that demand lengthy reasoning traces are particularly difficult for CoT-based methods. With only 1000 training examples, the ``Direct pred'' baseline---which employs an 8-layer Transformer identical in size to HRM---fails entirely on these challenging reasoning problems. When trained on the larger \textit{Sudoku-Extreme-Full} dataset, however, ``Direct pred'' can solve some easy Sudoku puzzles and reaches $16.9\%$ accuracy (see \Cref{fig:perf-layers}). \citet{searchformer2024} showed that a large vanilla Transformer model with 175M parameters, trained on 1 million examples across multiple trials, achieved only marginal success on 30x30 Maze tasks, with accuracy below $20\%$ using the $pass@64$ evaluation metric.


\subsection{Visualization of intermediate timesteps}

\begin{figure}[t]
  \centering
  \makebox[\linewidth][c]{%
    \includegraphics[width=1.1\linewidth]{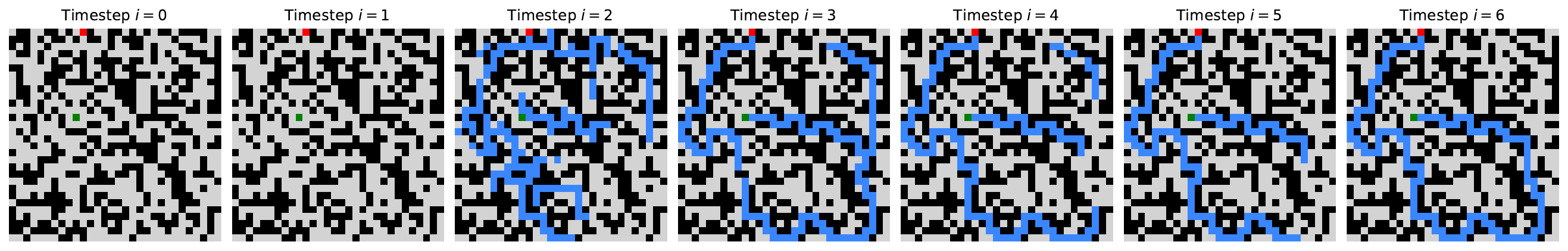}%
  }\\
  \makebox[\linewidth][c]{%
    \includegraphics[width=1.1\linewidth]{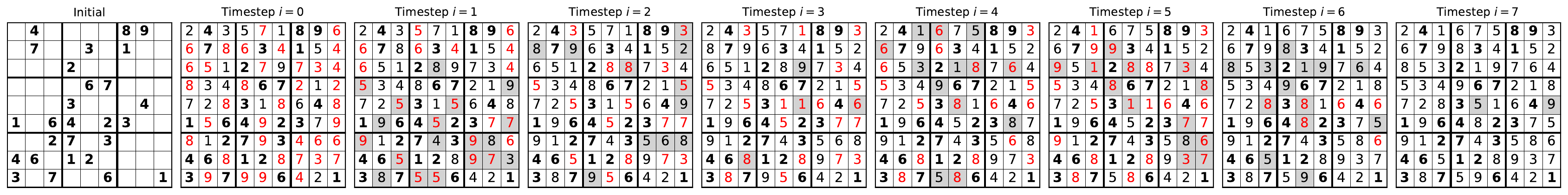}%
  }\\
  \makebox[\linewidth][c]{%
    \includegraphics[width=1.1\linewidth]{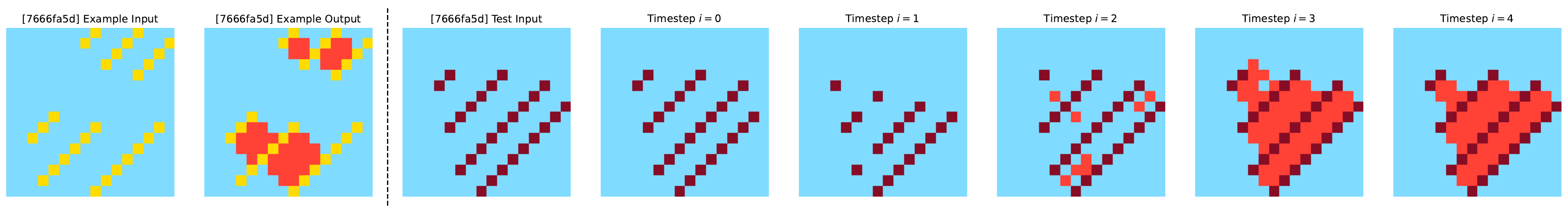}%
  }\\
  \makebox[\linewidth][c]{%
    \includegraphics[width=1.1\linewidth]{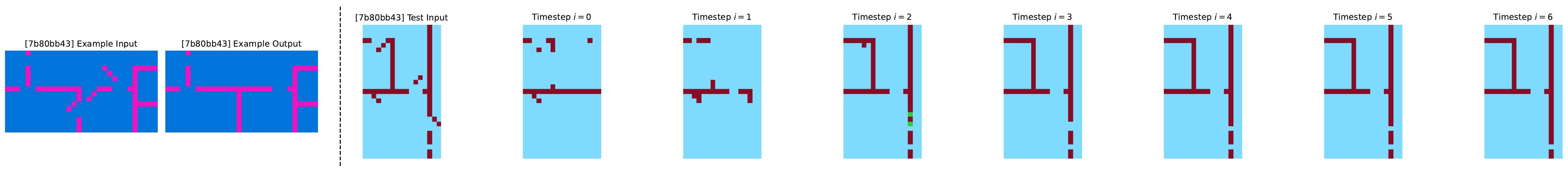}%
  }
  \caption{\textbf{Visualization of intermediate predictions by HRM on benchmark tasks.} \textbf{Top:} \textit{Maze-Hard}—blue cells indicate the predicted path. \textbf{Middle:} \textit{Sudoku-Extreme}—bold cells represent initial givens; red highlights cells violating Sudoku constraints; grey shading indicates changes from the previous timestep.
  \textbf{Bottom:} ARC-AGI-2 Task—left: provided example input-output pair; right: intermediate steps solving the test input.}
  \label{fig:visualization}
\end{figure}






Although HRM demonstrates strong performance on complex reasoning tasks, it raises an intriguing question: what underlying reasoning algorithms does the HRM neural network actually implement? Addressing this question is important for enhancing model interpretability and developing a deeper understanding of the HRM solution space. 

While a definitive answer lies beyond our current scope, we begin our investigation by analyzing state trajectories and their corresponding solution evolution. More specifically, at each timestep $i$ and given the low-level and high-level state pair ($z_L^i$ and $z_H^i$) we perform a preliminary forward pass through the H-module to obtain $\bar{z}^{i} = f_H(z_H^i, z_L^i; \theta_H)$ and its corresponding decoded prediction $\bar{y}^{i} = f_O(\bar{z}^{i}; \theta_O)$. The prediction $\bar{y}^{i}$ is then visualized in \Cref{fig:visualization}.   

In the Maze task, HRM appears to initially explore several potential paths simultaneously, subsequently eliminating blocked or inefficient routes, then constructing a preliminary solution outline followed by multiple refinement iterations. In Sudoku, the strategy resembles a depth-first search approach, where the model appears to explore potential solutions and backtracks when it hits dead ends. HRM uses a different approach for ARC tasks, making incremental adjustments to the board and iteratively improving it until reaching a solution. Unlike Sudoku, which involves frequent backtracking, the ARC solution path follows a more consistent progression similar to hill-climbing optimization. 
 
Importantly, the model shows that it can adapt to different reasoning approaches, likely choosing an effective strategy for each particular task. Further research is needed to gain more comprehensive insights into these solution strategies.

\section{Brain Correspondence}
\label{sec:brain_correspondence}

\begin{figure}[htbp]
\centering
\includegraphics[width=0.95\linewidth]{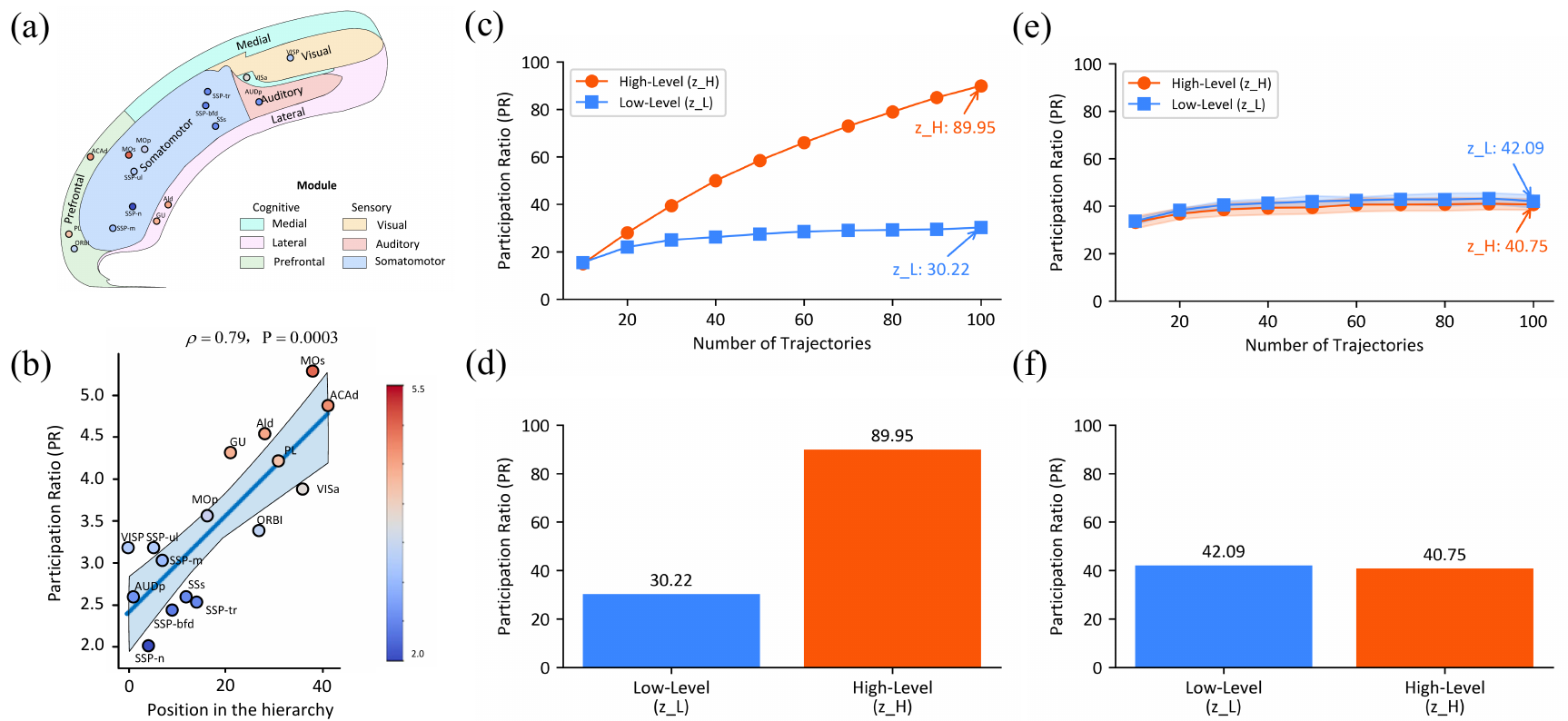}
\caption{
    \textbf{Hierarchical Dimensionality Organization in the HRM and Mouse Cortex.}
    (a,b) are adapted from~\citet{posani2025rarely}.
    (a) Anatomical illustration of mouse cortical areas, color-coded by functional modules.
    (b) Correlation between Participation Ratio (PR), a measure of effective neural dimensionality, and hierarchical position across different mouse cortical areas. Higher positions in the hierarchy (e.g., MOs, ACAd) exhibit significantly higher PR values compared to lower sensory areas (e.g., SSp-n), with a Spearman correlation coefficient of $\rho$ = 0.79 (P = 0.0003).
    (c,d) \textbf{Trained HRM.}
    (c) PR scaling of the trained HRM with task diversity. The dimensionality of the high-level module ($z_H$) scales with the number of unique tasks (trajectories) included in the analysis, indicating an adaptive expansion of its representational capacity. In contrast, the low-level module's ($z_L$) dimensionality remains stable.
    (d) PR values for the low-level ($z_L$, PR = 30.22) and high-level ($z_H$, PR = 89.95) modules of the \emph{trained} HRM, computed from neural activity during 100 unique Sudoku-solving trajectories. A clear dimensionality hierarchy is observed, with the high-level module operating in a substantially higher-dimensional space.
    (e,f) \textbf{Analysis of Untrained Network.} To verify that the dimensionality hierarchy is an emergent property of training, the same analyses were performed on an \emph{untrained} HRM with random weights.
    (e) In contrast to the trained model's scaling in (c), the dimensionality of both modules in the untrained model remains low and stable, failing to scale with the number of tasks.
    (f) Similarly, contrasting with the clear separation in (d), the PR values for the untrained model's modules ($z_L$, PR = 42.09; $z_H$, PR = 40.75) are low and nearly identical, showing no evidence of hierarchical separation. This confirms that the observed hierarchical organization of dimensionality is a learned property that emerges through training, not an artifact of the model's architecture.
}
\label{fig:hierarchical_dimensionality}
\end{figure}

A key principle from systems neuroscience is that a brain region's functional repertoire—its ability to handle diverse and complex tasks—is closely linked to the dimensionality of its neural representations~\citep{rigotti2013importance, mante2013context}. Higher-order cortical areas, responsible for complex reasoning and decision-making, must handle a wide variety of tasks, demanding more flexible and context-dependent processing~\citep{miller2001integrative}. In dynamical systems, this flexibility is often realized through higher-dimensional state-space trajectories, which allow for a richer repertoire of potential computations~\citep{maass2002realtime}. This principle gives rise to an observable \emph{dimensionality hierarchy}, where a region's position in the processing hierarchy correlates with its \emph{effective dimensionality}. To quantify this phenomenon, we can examine the Participation Ratio (PR), which serves as a standard measure of the effective dimensionality of a high-dimensional representation~\citep{altan2021estimating}. The PR is calculated using the formula 
\begin{equation*}
    \text{PR} = \frac{(\sum_{i} \lambda_i)^2}{\sum_{i} \lambda_i^2} \,,
\end{equation*}
where $\{\lambda_i\}$ are the eigenvalues of the covariance matrix of neural trajectories.
Intuitively, a higher PR value signifies that variance is distributed more evenly across many dimensions, corresponding to a higher-dimensional representation. Conversely, a lower PR value indicates that variance is concentrated in only a few principal components, reflecting a more compact, lower-dimensional structure.

The dimensionality hierarchy can be observed, for example, in the mouse cortex, where the PR of population activity increases monotonically from low-level sensory areas to high-level associative areas, supporting this link between dimensionality and functional complexity~\citep{posani2025rarely} (\Cref{fig:hierarchical_dimensionality} (a,b)).

We evaluated whether HRM reproduces this neuroscientific principle by calculating the PR for both recurrent modules after training on the \textit{Sudoku-Extreme Full} dataset. The PR computation used the covariance matrix derived from neural states gathered across multiple Sudoku-solving trajectories. The results show a striking parallel to the biological findings. The low-level module's state ($z_L$) occupies a relatively small subspace with a participation ratio of 30.22, whereas the high-level module's state ($z_H$) operates in a substantially larger subspace with a participation ratio of 89.95, as shown in \Cref{fig:hierarchical_dimensionality}(c).
Furthermore, \Cref{fig:hierarchical_dimensionality}(d) shows that increasing the number of unique tasks (trajectories) from 10 to 100 causes $z_H$ dimensionality to scale up accordingly, while $z_L$ dimensionality remains stable. These results suggest an \emph{emergent} separation of representational capacity between the modules that parallels their functional roles.

To confirm that this hierarchical organization is an emergent property of training, and not an artifact of the network's architecture, we performed a control analysis using an identical but untrained network with random weights.

We initialized an identical HRM architecture with random weights and, without any training, measured the PR of its modules as the network processed the same task-specific inputs given to the trained model.

The results, shown in
\Cref{fig:hierarchical_dimensionality}(e,f), reveal a stark contrast: the high-level and
low-level modules of the untrained network exhibit no hierarchical separation, with their PR values
remaining low and nearly indistinguishable from each other. This control analysis validates that
the dimensionality hierarchy is an \emph{emergent property} that arises as the model
learns to perform complex reasoning.

The high-to-low PR ratio in HRM ($z_H/z_L \approx 2.98$) closely matches that measured in the mouse
cortex ($\approx 2.25$). In contrast, conventional deep networks often exhibit \emph{neural collapse},
where last-layer features
converge to a low-dimensional
subspace~\citep{papyan2020prevalence,fang2021layerpeeled,zhu2021geometric}.
HRM therefore departs from the collapse pattern and instead fosters a high-dimensional representation in its higher module. This is significant because such representations are considered crucial for cognitive flexibility and are a hallmark of higher-order brain regions like the prefrontal cortex (PFC), which is central to complex reasoning.

This structural parallel suggests the model has discovered a fundamental organizational principle. By learning to partition its representations into a high-capacity, high-dimensional subspace ($z_H$) and a more specialized, low-dimensional one ($z_L$), HRM autonomously discovers an organizational principle that is thought to be fundamental for achieving robust and flexible reasoning in biological systems. This provides a potential mechanistic explanation for the model's success on complex, long-horizon tasks that are intractable for models lacking such a differentiated internal structure. We emphasize, however, that this evidence is correlational. While a causal link could be tested via intervention (e.g., by constraining the H-module's dimensionality), such methods are difficult to interpret in deep learning due to potential confounding effects on the training process itself. Thus, the causal necessity of this emergent hierarchy remains an important question for future investigation.

\section{Related Work}

\paragraph{Reasoning and algorithm learning}
Given the central role of reasoning problems and their close relation to algorithms, researchers have long explored neural architectures that enable algorithm learning from training instances. This line of work includes Neural Turing Machines (NTM)~\citep{NTK2014}, the Differentiable Neural Computer (DNC)~\citep{DNC2016}, and Neural GPUs~\citep{KS-2016}--all of which construct iterative neural architectures that mimic computational hardware for algorithm execution, and are trained to learn algorithms from data.
Another notable work in this area is Recurrent Relational Networks (RRN) \cite{Palm2017RecurrentRN}, which executes algorithms on graph representations through graph neural networks.

Recent studies have integrated algorithm learning approaches with Transformer-based architectures. 
Universal Transformers extend the standard Transformer model by introducing a recurrent loop over the layers and implementing an adaptive halting mechanism. \citet{geiping2025} demonstrate that looped Transformers can generalize to a larger number of recurrent steps during inference than what they were trained on. \citet{CoconutLatentReasoning2024} propose adding continuous recurrent reasoning tokens to the Transformer. Finally, TransNAR \cite{Bounsi2024TransformersMN} combine recurrent graph neural networks with language models.

Building on the success of CoT-based reasoning, a line of work have introduced fine-tuning methods that use reasoning paths from search algorithms (like A*) as SFT targets~\cite{Liu2023GoatFL, searchformer2024, dualformer2025}.

We also mention adaptive halting mechanisms designed to allocate additional computational resources to more challenging problems. This includes the Adaptive Computation Time (ACT) for RNNs~\cite{Graves2016AdaptiveCT} and follow-up research like PonderNet~\cite{Banino2021PonderNetLT}, which aims to improve the stability of this allocation process.

HRM further pushes the boundary of algorithm learning through a brain-inspired computational architecture that achieves exceptional data efficiency and model expressiveness, successfully discovering complex and diverse algorithms from just 1000 training examples.

\paragraph{Brain-inspired reasoning architectures}

Developing a model with the reasoning power of the brain has long been a goal in brain-inspired computing.
Spaun~\citep{spaun} is one notable example, which uses spiking neural networks to create distinct modules corresponding to brain regions like the visual cortex and prefrontal cortex. This design enables an architecture to perform a range of cognitive tasks, from memory recall to simple reasoning puzzles. However, its reasoning relies on hand-designed algorithms, which may limit its ability to learn new tasks.
Another significant model is the Tolman-Eichenbaum Machine (TEM)~\citep{TEM}, which is inspired by the hippocampal-entorhinal system's role in spatial and relational memory tasks. TEM proposes that medial entorhinal cells create a basis for structural knowledge, while hippocampal cells link this basis to sensory information. This allows TEM to generalize and explains the emergence of various cell types like grid, border, and place cells. 
Another approach involves neural sampling models~\citep{buesing2011neural}, which view the neural signaling process as inference over a distribution, functioning similarly to a Boltzmann machine. These models often require hand-made rules to be set up for solving a specific reasoning task.
In essence, while prior models are restricted to simple reasoning problems, HRM is designed to solve complex tasks that are hard for even advanced LLMs, without pre-training or task-specific manual design.

\paragraph{Hierarchical memory}

The hierarchical multi-timescale structure also plays an important role in how the brain processes memory. Models such as Hierarchical Sequential Models~\cite{NIPS1995_c667d53a} and Clockwork RNN~\cite{Koutnk2014ACR} use multiple recurrent modules that operate at varying time scales to more effectively capture long-range dependencies within sequences, thereby mitigating the forgetting issue in RNNs. 

Similar mechanisms have also been adopted in linear attention methods for memorizing long contexts (see the Discussions section). 
Since HRM focuses on reasoning, full attention is applied for simplicity. Incorporating hierarchical memory into HRM could be a promising future direction.

\section{Discussions}

\paragraph{Turing-completeness of HRM}

Like earlier neural reasoning algorithms including the Universal Transformer~\citep{UniversalTransformer2018}, HRM is computationally universal when given sufficient memory and time constraints. In other words, it falls into the category of models that can simulate any Turing machine, overcoming the computational limitations of standard Transformers discussed previously in the introduction. Given that earlier neural algorithm reasoners were trained as recurrent neural networks, they suffer from premature convergence and memory intensive BPTT. Therefore, in practice, their effective computational depth remains limited, though still deeper than that of a standard Transformer. By resolving these two challenges and being equipped with adaptive computation, HRM could be trained on long reasoning processes, solve complex puzzles requiring intensive depth-first search and backtracking, and move closer to  practical Turing-completeness.

\paragraph{Reinforcement learning with chain-of-thought}

Beyond fine-tuning using human-annotated CoT, reinforcement learning (RL) represents another widely adopted training methodology. 
However, recent evidence suggests that RL primarily unlocks existing CoT-like capabilities rather than discovering fundamentally new reasoning mechanisms~\citep{wang2025,Muennighoff2025s1, lightr12025, limr2025}.
Additionally, CoT-training with RL is known for its instability and data inefficiency, often requiring extensive exploration and careful reward design. In contrast, HRM takes feedback from dense gradient-based supervision rather than relying on a sparse reward signal. Moreover, HRM 
operates naturally in a continuous space, which is biologically plausible and avoids allocating same computational resources to each token, even though tokens vary in their reasoning and planning complexity~\cite{CoconutLatentReasoning2024}.

\paragraph{Linear attention}

Recurrence has been explored not only for its capability in universal computation, but also as a means to replace the attention mechanism in Transformers, which suffers from quadratic time and memory complexity~\citep{Dao2024TransformersAS}. Recurrent alternatives offer a more efficient design by processing input tokens sequentially and predicting the next token at each time step, similar to early RNN-based language models.

Some linear-attention variants, such as Log-linear Attention \cite{guo2025log}, share an RNN-like state-update that can be interpreted as propagating multi-timescale summary statistics, thereby retaining long-range context without the quadratic memory growth of standard self-attention.
However, substituting the attention mechanism alone does not change the fact that Transformers are still fixed-depth, and require CoT as a compensatory mechanism. Notably, linear attention can operate with a reduced key-value cache over extended contexts, making them more suitable for deployment on resource-constrained edge devices.

\section{Conclusion}

This work introduces the Hierarchical Reasoning Model, a brain-inspired architecture that leverages hierarchical structure and multi-timescale processing to achieve substantial computational depth without sacrificing training stability or efficiency. With only 27M parameters and training on just 1000 examples, HRM effectively solves challenging reasoning problems such as ARC, Sudoku, and complex maze navigation--tasks that typically pose significant difficulties for contemporary LLM and chain-of-thought models.

Although the brain relies heavily on hierarchical structures to enable most cognitive processes, these concepts have largely remained confined to academic literature rather than being translated into practical applications. 
The prevailing AI approach continues to favor non-hierarchical models. Our results challenge this established paradigm and suggest that the Hierarchical Reasoning Model represents a viable alternative to the currently dominant chain-of-thought reasoning methods, advancing toward a foundational framework capable of Turing-complete universal computation.

\textbf{Acknowledgements} We thank Mingli Yuan, Ahmed Murtadha Hasan Mahyoub and Hengshuai Yao for their insightful discussions and valuable feedback throughout the course of this work.

\clearpage
\begin{hyphenrules}{nohyphenation}
\setlength{\bibsep}{.5ex plus .8ex}
\bibliographystyle{unsrtnat}
\bibliography{main}
\end{hyphenrules}

\clearpage
\appendix

\end{document}